\documentclass[conference]{IEEEtran}
\usepackage{cite}
\usepackage{rotating}
\usepackage{graphicx}
\usepackage{rotating}
\usepackage{amsmath}
\usepackage{algorithmic}
\usepackage{color}
\usepackage{xspace}
\usepackage{array}
\usepackage{hyperref}
\usepackage{float}
\ifCLASSOPTIONcompsoc
  \usepackage[caption=false,font=normalsize,labelfont=sf,textfont=sf]{subfig}
\else
  \usepackage[caption=false,font=footnotesize]{subfig}
\fi
\usepackage{stfloats}
\usepackage{url}

\usepackage{flushend}

\newcommand{\comment}[1]{}

\usepackage[dvipsnames]{xcolor}

\begin{document}
\title{Runway Sign Classifier: A DAL~C Certifiable Machine Learning System}

\makeatletter 
\newcommand{\linebreakand}{%
  \end{@IEEEauthorhalign}
  \hfill\mbox{}\par
  \mbox{}\hfill\begin{@IEEEauthorhalign}
}
\makeatother

\author{
\IEEEauthorblockN{Konstantin Dmitriev}
\IEEEauthorblockA{Technical University of Munich\\
Garching, Germany\\
konstantin.dmitriev@tum.de}
\and
\IEEEauthorblockN{Johann Schumann}
\IEEEauthorblockA{
KBR, NASA Ames Research Center\\
Moffett Field, CA\\
johann.m.schumann@nasa.gov}
\and
\IEEEauthorblockN{Islam Bostanov}
\IEEEauthorblockA{Technical University of Munich\\
Garching, Germany\\
islam.bostanov@tum.de}
\linebreakand
\IEEEauthorblockN{Mostafa Abdelhamid}
\IEEEauthorblockA{Technical University of Munich\\
Garching, Germany\\
mostafa.abdelhamid@tum.de}
\and
\IEEEauthorblockN{Florian Holzapfel}
\IEEEauthorblockA{
Technical University of Munich\\
Garching, Germany\\
florian.holzapfel@tum.de}
}

\maketitle
\graphicspath{{../figures/}}

\makeatletter
\def\ps@IEEEtitlepagestyle{%
\def\@oddfoot{\mycopyrightnotice}%
\def\@evenfoot{}%
}

\def\mycopyrightnotice{%
{\footnotesize
\vbox{%
  \hbox{\copyright~2023 IEEE. This material was accepted for publication at the 2023 IEEE/AIAA 42nd Digital Avionics Systems Conference (DASC 2023). Personal use}%
  \hbox{ of this material is permitted. Permission from IEEE must be obtained for all other uses, in any current or future media, including reprinting/republishing}
  \hbox{ this material for advertising or promotional purposes, creating new collective works, for resale or redistribution to servers or lists, or reuse of any}%
  \hbox{ copyrighted component of this work in other works.}%
}
}
\gdef\mycopyrightnotice{}
}

\begin{abstract}
In recent years, the remarkable progress of Machine Learning (ML) technologies within the domain of Artificial Intelligence (AI) systems has presented unprecedented opportunities for the aviation industry, paving the way for further advancements in automation, including the potential for single pilot or fully autonomous operation of large commercial airplanes. However, ML technology faces major incompatibilities with existing airborne certification standards, such as  ML model traceability and explainability issues or the inadequacy of traditional coverage metrics. Certification of ML-based airborne systems using current standards is problematic due to these challenges. This paper presents a case study of an airborne system utilizing a Deep Neural Network (DNN) for airport sign detection and classification. Building upon our previous work, which demonstrates compliance with Design Assurance Level (DAL) "D", we upgrade the system to meet the more stringent requirements of Design Assurance Level "C". To achieve DAL~C, we employ an established architectural mitigation technique involving two redundant and dissimilar Deep Neural Networks. The application of novel ML-specific data management techniques further enhances this approach. This work is intended to illustrate how the certification challenges of ML-based systems can be addressed for medium criticality airborne applications.
\end{abstract}

\section{Introduction and Related Work}
\label{sec: intro}
The remarkable progress of Machine Learning (ML) technologies in recent years has the potential to revolutionize aviation \cite{easaroadmap2-2023}. Data-driven ML systems can implement highly complex cognitive functions such as vision and language processing that can enable single pilot or fully autonomous operation of large commercial airplanes, a level of automation not possible with traditional rule-based software systems \cite{codann_1_2020, codann_2_2020}. However, ML technology encounters various inherent incompatibilities with existing airborne certification standards. These incompatibilities encompass challenges related to explainability, traceability, and implementation coverage \cite{eurocae2021SoC}. As a result, the utilization of ML-based applications within the framework of existing airborne certification standards is currently impeded.

The industry, aviation authorities, and academia are actively engaged in collaborative efforts to develop new certification standards aimed at addressing the existing incompatibilities of ML technology with existing certification practices. The development of a new standard for airborne machine learning (ML) certification has been underway since 2019 through the collaborative efforts of the EUROCAE/SAE WG-114\footnote{\url{https://www.eurocae.net/news/posts/2019/june/new-working-group-wg-114-artificial-intelligence/}}/G-34\footnote{\url{https://standardsworks.sae.org/standards-committees/g-34-artificial-intelligence-aviation}} joint working group. The working group has published reports reflecting the intermediate results \cite{eurocae2021SoC, wg114_ml_cert}. The European Aviation Safety Agency (EASA) has released the guidance for ML applications [8], which includes anticipated objectives and means of compliance for the certification of safety-critical airborne systems based on machine learning. In collaboration with Daedalean AG, the U.S. Federal Aviation Administration (FAA) has released a research report on a neural network vision-based landing guidance system \cite{FAA_ML_Study}. This report offers a practical assessment of the previously proposed W-shaped process for ML applications, as outlined in earlier Daedalean AG publications \cite{codann_1_2020,codann_2_2020}. Numerous ongoing research projects in academia are exploring various aspects of ML-based systems certification in different domains \cite{deel_ML,torens2023_ml_cert,escudero_ml_visual_land_cert,feather2023assurance,abeywickrama2023aeros}.

However, all currently available materials do not constitute final certification standards but rather anticipated guidance specific to systems utilizing machine learning technology. Due to the novelty and complexity of the subject matter, as well as the involvement of multiple stakeholders, the accomplishment and acceptance of new certification standards for ML may require several additional years. The absence of certification standards presents a fundamental constraint to the effective utilization of the advantages offered by ML technology within the aviation industry.
To tackle these challenges, we proposed and implemented in our previous works \cite{ML_DALD_Dmitriev_21, ML_DAL_D_case_study} a custom ML certification workflow for low-criticality (Design Assurance Level ”D”)\cite{SAE.4754} machine learning systems which is based on existing certification standards. In this work, we present an extension to our previous case study \cite{ML_DAL_D_case_study} of the Machine Learning Runway Sign Classifier (RSC) system designed to detect and classify airport runway signs utilizing a computer vision deep neural network. Through this work we illustrate the implementation of the next version of our custom workflow addressing DAL~C certification objectives as proposed in \cite{ML_DAL_C_workflow}. To achieve DAL~C, we utilize the proven architectural mitigation approach in combination with novel ML-specific development and verification techniques. This work is intended to demonstrate the incremental certification approach from low DAL~D to medium DAL~C assurance level by the application of existing assurance practices.

The rest of this paper is structured as follows: Section~\ref{sec:case-study} provides an overview of the RSC system under study including a discussion of its use cases, an overview of the system requirements and architecture, data management aspects, and an analysis of system component implementation, integration, verification and validation. Section~\ref{sec:cert_analysis} comprises an evaluation of the RSC system compliance with the DAL~C certification objectives. Section~\ref{sec:conclusions} provides a summary of the paper, explores future work, and presents the concluding remarks.

\section{Case Study}
\label{sec:case-study} 
In this work, we upgrade the ML-based Runway Sign Classifier system introduced in \cite{ML_DAL_D_case_study} to achieve airborne Design Assurance Level C. RSC system relies on machine-learning DNN technology which constitutes the focal point of our research and requires special certification considerations that we studied in \cite{ML_DAL_C_workflow} and implemented as a realistic example in this work. In order to meet the DAL~C certification objectives for RSC, we utilize architectural mitigation along with ML-specific assurance techniques, following the DAL~C ML workflow proposed in \cite{ML_DAL_C_workflow}.

\begin{figure*}[!htb]
	\centering
	\includegraphics[width=0.8\textwidth]{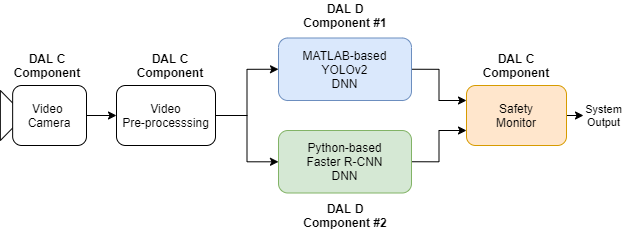}
	\caption{Runway Sign Classifier DAL~C System Architecture}
	\label{fig:sys_arch}
\end{figure*}

\begin{figure}[!hb]
	\centering
	\includegraphics[width=0.6\columnwidth]{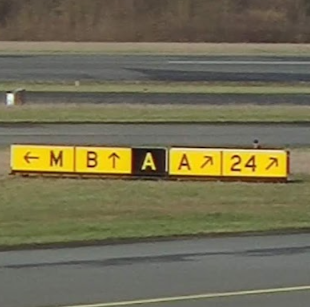}
	\caption{Airport sign}
	\label{fig:airport_sign}
\end{figure}

\subsection{System Overview}
Runway Sign Classifier (RSC) system is intended for the detection and classification of airport signs (Fig.~\ref{fig:airport_sign}). The system receives visual images from a forward-facing camera installed on the aircraft and uses a DNN for detection and classification tasks. In our baseline case study \cite{ML_DAL_D_case_study}, we considered three variants of such a system depending on the level of automation and corresponding criticality levels:
\begin{itemize}
    \item RSC-FB (Flight Bag). This variant serves as a pilot assistance system by providing information about detected runway signs to pilots e.g., via an enhanced vision system. Failures of such a system would have minor safety effects, as the pilot is still responsible for the continuous perception of the visual environment and retains full control of the aircraft. Consequently, the RSC-FB can be assigned the lowest design assurance level (DAL~D).
    \item RSC-SM (Safety Monitor). This system variant provides information about detected airport signs to enable cockpit annunciation or automatic braking in hazardous taxing situations, such as crossing a holding position sign without the controller's approval. This capability provides a higher autonomy and criticality level 
    and can be assigned DAL~C.
    \item RSC-AU (Autonomy). This RSC variant is intended to support fully autonomous taxiing. In this context, the location and sign data directly contribute to the decision-making system responsible for autonomous aircraft control and preventing runway incursions. Given the potential safety impact, the system can be assigned DAL~B or DAL~A, depending on the aircraft category.
\end{itemize}
In our previous work \cite{ML_DAL_D_case_study}, we implemented the first variant of the RSC-FB system and demonstrated its compliance with DAL~D certification objectives. This study focuses on implementing the second variant, RSC-CM, using our custom ML workflow for DAL~C \cite{ML_DAL_C_workflow}. In future work, we plan to address the third and most critical variant, RSC-AU.

\subsection{System Architecture}
\label{sec:sys_arch_reqs}

A single-channel architecture of the original Runway Sign Classifier system presented in \cite{ML_DAL_D_case_study} is compatible with the DAL~D custom ML workflow \cite{ML_DALD_Dmitriev_21} but has to be redesigned to implement the architectural mitigation approach proposed DAL~C ML-based systems in \cite{ML_DAL_C_workflow}. We upgraded the RSC system architecture by adding a second dissimilar DNN and a safety monitor component. Fig.~\ref{fig:sys_arch} shows the implemented in this work DAL~C RSC system architecture with two independent DNNs running in parallel; outputs of DNNs are continuously monitored by the safety monitor component and inhibited in case of divergence above the specified threshold. As justified in \cite{ML_DAL_C_workflow}, the design assurance levels of the redundant dissimilar DNN components can be reduced to DAL~D in such architecture while maintaining the DAL~C for the overall system.



\subsection{RSC Requirements}
The functional and operational requirements that we developed for the DAL~D implementation of the RSC system in our previous study \cite{ML_DAL_D_case_study} remain unchanged for the DAL~C architecture upgrade, therefore we reused them in this case study. To address the DAL~C workflow objectives for architectural mitigation measures \cite{ML_DAL_C_workflow}, we added DAL~C specific requirements for system components dissimilarity and a safety monitor as detailed in  Table~\ref{tab:arch_req}. Furthermore, we elaborated the RSC operational conditions requirements into explicit requirements for RSC datasets (Table~\ref{tab:data_req}) with the purpose of supporting data-specific verification activities discussed in Section~\ref{sec:data}. For requirements authoring and management, we utilized the Polarion life-cycle management framework \cite{polarion_2021_schmiechen}.

\begin{table}[htb]
\begin{tabular}{|p{0.12\columnwidth}|p{0.77\columnwidth}|}
\hline
ID & Description \\
\hline
\hline
RSC-A1&
The RSC system shall include two dissimilar DNN components independently performing sign detection function\\
\hline
RSC-A2&
The RSC system shall include a safety monitor component that compares the outputs of the dissimilar DNN components. The safety monitor shall inhibit the system outputs if the outputs of the DNN components do not fall within the specified tolerance\\
\hline
RSC-A3&
DNN components shall utilize dissimilar network architecture\\
\hline
RSC-A4&
The RSC DNN components shall be trained and tested using independent datasets\\
\hline
RSC-A5&
The RSC DNN components software shall be implemented using different programming languages\\
\hline
RSC-A6&
The RSC DNN components shall be implemented using dissimilar electronic hardware\\
\hline
RSC-A7&
The RSC DNN components shall be implemented by different individuals\\
\hline
RSC-A8&
Verification of the RSC DNN components shall be performed by different individuals or groups\\
\hline
\end{tabular}

\caption{RSC system architecture requirements}
\label{tab:arch_req}
\end{table}

\begin{table}[htb]
\begin{tabular}{|p{0.10\columnwidth}|p{0.80\columnwidth}|}
\hline
ID & Description \\
\hline
\hline
\multicolumn{2}{|p{0.9\columnwidth}|}{
{
The RSC datasets shall contain sign images that have been captured within the premises of the following airports:}}\\
\hline
KSFO &
San Francisco International Airport\\
\hline
KBOS &
Boston Logan International Airport\\
\hline
KSAN &
San Diego International Airport\\
\hline
\multicolumn{2}{|p{0.9\columnwidth}|}{
The RSC datasets shall contain sign images that have been captured under the following weather conditions:}\\
\hline
FAIR &
Fair weather\\
\hline
RAIN &
Rainy weather\\
\hline
SNOW &
Snowy weather\\
\hline
FOG &
Foggy weather\\
\hline
\multicolumn{2}{|p{0.9\columnwidth}|}{
The RSC datasets shall contain sign images that have been captured during the following time of day:}\\
\hline
MRNG &
Morning\\
\hline
DUSK &
Dusk\\
\hline
AFTN &
Afternoon\\
\hline
DAWN &
Dawn\\
\hline
\multicolumn{2}{|p{0.9\columnwidth}|}{
The RSC datasets shall contain sign images that have been captured within the following distance ranges:}\\
\hline
DS10 &
10 to 12 meters\\
\hline
DS12 &
12 to 14 meters\\
\hline
DS14 &
14 to 16 meters\\
\hline
\multicolumn{2}{|p{0.9\columnwidth}|}{
The RSC datasets shall contain sign images captured from points at an elevation above ground level within the following ranges:}\\
\hline
EL10 &
1.0 to 1.3 meters\\
\hline
EL13 &
1.3 to 1.6 meters\\
\hline
EL16 &
1.6 to 1.9 meters\\
\hline
\multicolumn{2}{|p{0.9\columnwidth}|}{
The RSC datasets shall contain sign images captured within the following ranges of the camera's lateral offset from the sign center line:}\\
\hline
LO00 &
0 to 0.7 meters\\
\hline
LO07 &
0.7 to 1.4 meters\\
\hline
LO14 &
1.4 to 2 meters\\
\hline
\end{tabular}
\caption{RSC Data Requirements}
\label{tab:data_req}
\end{table}

\subsection{ML Data Management}
\label{sec:data}
In ML development workflows, datasets play a crucial role as they directly define the system behavior and therefore serve as high-level software requirements \cite{ML_DAL_D_case_study}. In this case study,  we address the independence aspects of datasets used for the different DNN components to satisfy the dissimilarity requirements outlined in Section~\ref{sec:sys_arch_reqs}. Additionally, we study the specific aspects of data configuration management and data verification that have to be addressed in a safety-critical context but are not covered well by the existing assurance practices.

\subsubsection{Datasets Independence}
As discussed in \cite{ML_DAL_C_workflow}, a significant factor that can contribute to common errors in DNN components is the use of a shared dataset for training and testing across different DNNs. To address this aspect, we developed a separate dataset for the Faster R-CNN DNN component by utilizing the X-Plane\footnote{\url{https://www.x-plane.com/}} flight simulator to generate airport visual scenes. X-Plane is one of the most advanced flight simulators that offers exceptional capabilities in generating photo-realistic scenery and is also used for scenario-based pilot training \cite{xplane2011williams}. For the YOLOv2 DNN component, we reused the dataset generated using the FlightGear flight simulator in the scope of our previous work \cite{ML_DAL_D_case_study}. To ensure data integrity and prevent inadvertent data leakage, we implemented separate Git repositories for managing the FlightGear and X-Plane datasets. This segregation ensures that each dataset remains independent.

\subsubsection{Data Configuration Management}
ML datasets are typically characterized by a large number of elements, typically images, and videos in compressed binary format. Traditional configuration management systems such as Git\footnote{\url{https://git-scm.com/}} or SVN\footnote{\url{https://subversion.apache.org/}} can experience performance issues and other limitations when tracking the configuration of binary files \cite{ML_config_mgmt_JANARDHANAN2020}. To address these issues in the scope of our case study, we used the DVC\footnote{\url{https://dvc.org/}} extension to the Git system to manage the configuration of our Runway Sign Classifier datasets. DVC is an open-source tool designed for machine learning and data-science applications. DVC enables storage of the binary data on a remote server allowing Git to manage a textual file with tracking reference of binary files. 

\subsection{System Implementation and Integration}

In this work, we focus on the implementation of the dissimilar Python-based R-CNN DNN component, the safety monitor component, and their integration with the YOLOv2 DNN component implemented in our previous case study \cite{ML_DAL_D_case_study}. YOLOv2 (Figure~\ref{fig:yolo2_arch}) is a widely-used object detection network known for its speed and relatively small size \cite{redmon2017yolo9000}. We used the MATLAB implementation of the YOLOv2 DNN \verb|yolov2ObjectDetector|\footnote{\url{https://www.mathworks.com/help/vision/ref/yolov2objectdetector.html}} utilizing the DarkNet-19 backbone network and conducting the YOLOv2 training employing the capabilities of the MATLAB Computer Vision Toolbox\footnote{\url{https://www.mathworks.com/help/vision/index.html}}.

\begin{figure*}[!htb]
	\centering
	\includegraphics[width=0.85\textwidth]{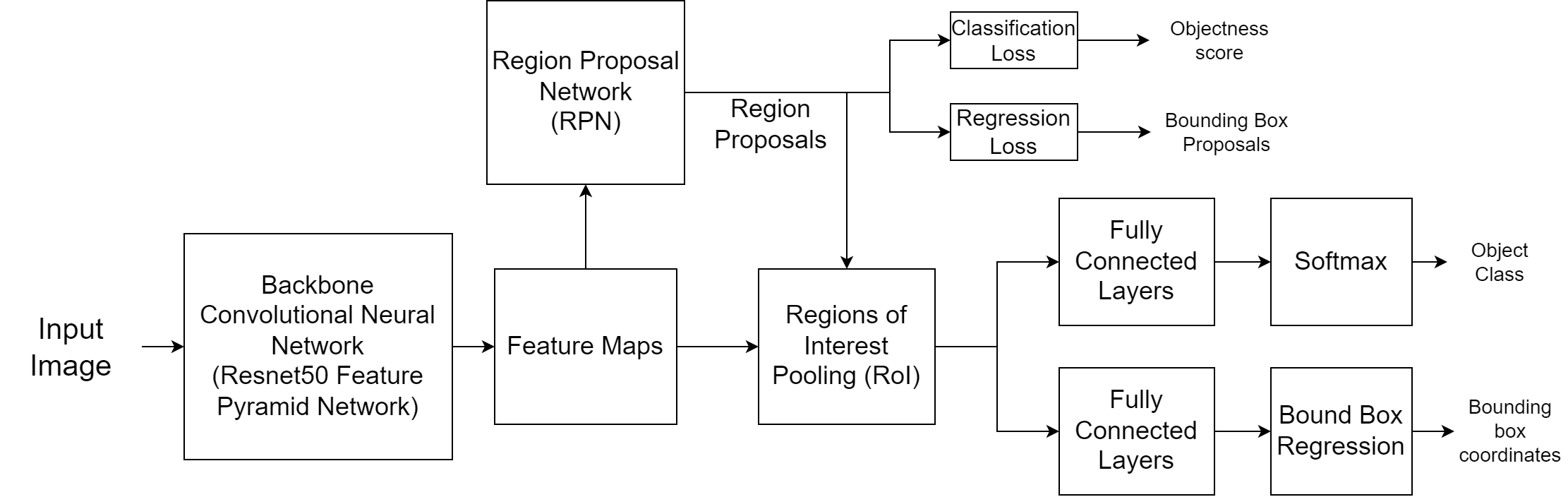}
	\caption{Faster R-CNN Architecture}
	\label{fig:f_rcnn_arch}
\end{figure*}

\begin{figure}[!htb]
	\centering
	\includegraphics[width=0.9\columnwidth]{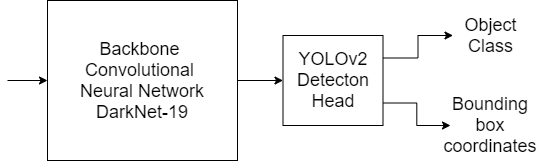}
	\caption{YOLOv2 DNN Architecture}
	\label{fig:yolo2_arch}
\end{figure}

The implementation and integration of the video camera and video pre-processing component are not included in this study because these components can be implemented and certified using traditional non-ML software and hardware assurance practices that fall outside the scope of this research.

\subsubsection{DNN components} The YOLOv2 DNN, implemented in our prior work \cite{ML_DAL_D_case_study}, was reused for the first RSC DNN component. Taking into account the dissimilarity requirements for DNN components provided in Section~\ref{sec:sys_arch_reqs}, we opted for a Python-based implementation of the Faster R-CNN \cite{rcnn2015ren} as a DNN architecture (Figure~\ref{fig:f_rcnn_arch}) for the second RSC component. The Faster R-CNN architecture employs a two-stage detection approach, incorporating a Region Proposal Network (RPN) to predict potential object locations before performing main network predictions and classification. The Faster R-CNN DNN architecture (depicted in Figure~\ref{fig:f_rcnn_arch}) presents notable differences compared to the YOLOv2 architecture (shown in Figure~\ref{fig:yolo2_arch}). These differences are evident in the backbone network architecture, with Faster R-CNN employing a ResNet50 convolutional network with 50 layers, while YOLOv2 uses a DarkNet-19 network with 19 layers. Moreover, the overall detection approach differs, with Faster R-CNN utilizing an advanced two-stage prediction pipeline, whereas YOLOv2 employs a single-stage pipeline optimized for speed.

We implemented the Faster R-CNN DNN component using the \verb|fasterrcnn_resnet50_fpn| model from the \verb|torchvision|\footnote{\url{https://pytorch.org/vision/main/index.html}} Python package. Utilizing the dissimilar architecture of DNN models, different programming languages and diverse training frameworks for independent DNN components mitigates the risk of common errors in independent system channels. Furthermore, to mitigate the risk of human error, we assigned the development of the Faster R-CNN DNN and YOLOv2 components to different individuals.

To match the interfaces of the YOLOv2 DNN component \cite{ML_DAL_D_case_study}, we adjusted the input and output layers parameters of Faster R-CNN DNN. We kept the default values of the other DNN parameters, such as the number of RPN input features and loss function balancing parameters, because the required component performance was achieved during training without other hyperparameters tuning.

The Faster R-CNN training process utilized the independent X-Plane dataset described in Section~\ref{sec:data}. To handle the training dataset, we employed the \verb|torch.transforms|\footnote{\url{https://pytorch.org/vision/main/transforms.html}} and \verb|CV2|\footnote{\url{https://pypi.org/project/opencv-python/}} Python packages for various tasks, including the conversion of dataset images into PyTorch tensors to provide compatibility with the model training framework.

To determine the DNN detection thresholds, we evaluated the trained Faster R-CNN DNN using a precision/recall metric \cite{precision_recall_2015_saito}. With a selected confidence and intersection threshold of 95\% delivering the required average precision and highest possible recall of the model, the final Faster R-CNN DNN achieved an average detection precision of 100\% on the test dataset. 

\subsubsection{Safety monitor component}

The safety monitor component implements protection against DNN failures that can result in incorrect system output. The safety monitor compares the outputs (bounding box coordinates and sign class) of the dissimilar YOLOv2 and Faster R-CNN DNNs and passes through the output of the first DNN if it matches the outputs of the second DNN. If the different DNN components do not return the same sign class or the divergence between bounding box coordinates is above the threshold, the safety monitor inhibits the system output by setting the signal validity flag to FALSE. We selected the Intersection over Union (IoU) metric \cite{iou2019rezatofighi} for measuring the coherence of the bounding boxes coordinates between the different DNN components. The safety monitor IoU is calculated by dividing the area of intersection between the regions predicted by different DNN components by the area of their union (Figure~\ref{fig:IoU_def}).

\begin{figure}[!htb]
	\centering
	\includegraphics[width=0.85\columnwidth]{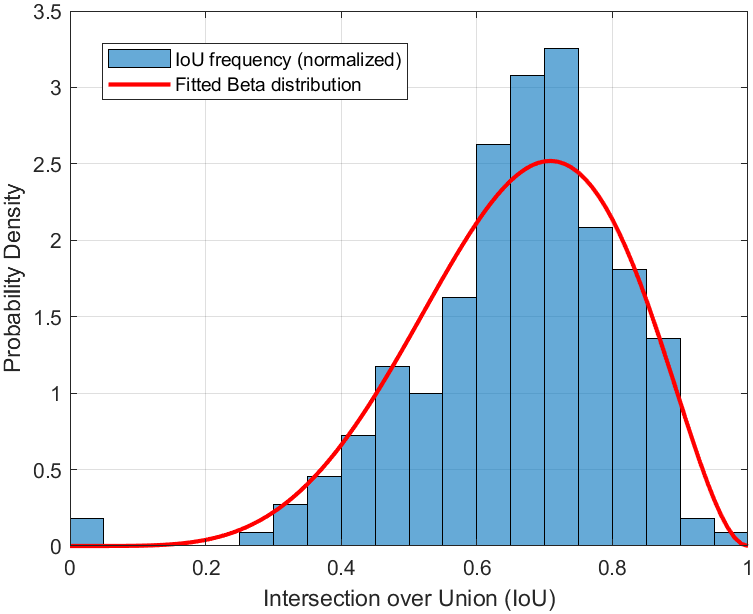}
	\caption{Distribution of IoU for dissimilar DNN components}
	\label{fig:IoU_dist}
\end{figure}

\begin{figure}[!htb]
	\centering
	\includegraphics[width=0.80\columnwidth]{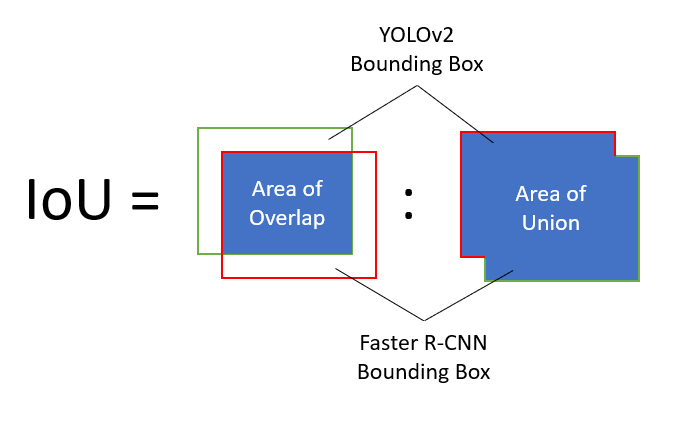}
	\caption{Intersection over Union}
	\label{fig:IoU_def}
\end{figure}

The IoU threshold for the safety monitor component was determined experimentally by analysis of the IoU distribution over the available test dataset. The experimental frequencies of the IoU values fit the Beta distribution ($\alpha=5.88, \beta=3.01$) with a mean of 0.66 and a standard deviation of 0.54. Figure~\ref{fig:IoU_dist} shows the observed IoU value frequencies on the FlightGear testing dataset and the probability density of the fitted Beta distribution. To ensure the system availability above 95\%, the safety monitor threshold for the IoU value was set to 0.32.


\subsection{ML Specific Verification}
\label{sec:vnv}
Verification activities serve the purpose of identifying potential errors introduced during development. In this case study, we focus on the ML-specific data verification techniques which complement testing, reviews, and analysis methods that we studied in our previous work to fulfill the verification objectives required for DAL D systems \cite{ML_DAL_D_case_study}.

\begin{figure*}[!htb]
\centering\includegraphics[width=0.8\textwidth]{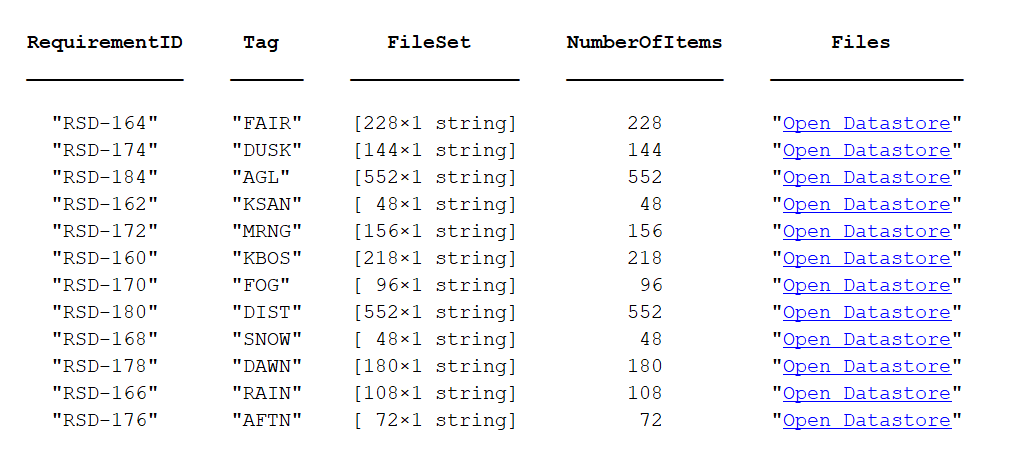} 
        \caption{Data Traceability Matrix} 
        \label{fig:trace_matrix}
\end{figure*}

\begin{figure}[!htb]
\centering\includegraphics[width=0.35\textwidth]{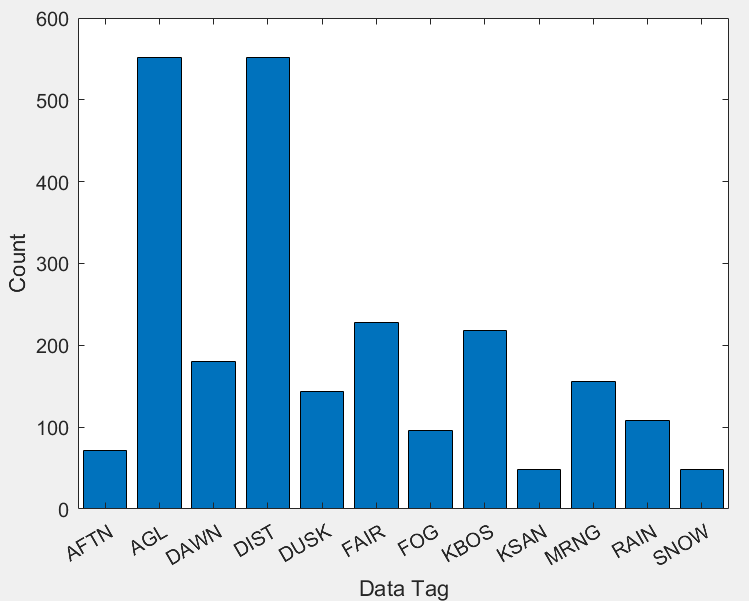} 
        \caption{Data Requirements Coverage Histogram} 
        \label{fig:data_req_cov_chart}
\end{figure}

\subsubsection{Data Verification}
ML datasets representing high-level software requirements in our workflow entail the application of typical objectives for requirements verification, such as verification of accuracy, consistency, and compliance with upstream requirements. However, ML datasets possess specific properties, such as higher complexity and size, that are not explicitly addressed in the current requirements verification practices outlined in DO-178C \cite{RTCA.178} and ARP-4754 \cite{SAE.4754}. In the scope of this case study, we implement a data traceability approach to address these specific properties of ML datasets. This approach is based on creating bi-directional links between textual data requirements and DNN datasets by tagging data files during the data preparation process. 

To collect and analyze traceability information for datasets, we developed the DataTrace extension for the SimPol traceability management tool \cite{hochstrasser2020mbd_simpol}, originally created for model-based design workflows. DataTrace automates the fetching of requirements from the requirements management system Polarion  \cite{polarion_2021_schmiechen}
and creates and stores links to the linked datasets. 

DataTrace also provides a capability to generate a data traceability matrix (Fig.~\ref{fig:trace_matrix}) and a data requirements coverage histogram (Fig.~\ref{fig:data_req_cov_chart}) from the collected data traceability information. We used the data traceability matrix to review by sampling the compliance of dataset elements with data requirements and analyze the traceability of dataset elements to requirements. The data requirement coverage histogram (Fig.~\ref{fig:data_req_cov_chart}) 
was used to analyze the coverage of data requirements, including the characteristics of coverage distribution.

\section{Certification Compliance Analysis}
\label{sec:cert_analysis}
In this section, we elaborate on the analysis of DO-178C DAL~C objectives provided in \cite{ML_DAL_C_workflow} by mapping the objectives to the actual activities performed and artifacts created in the scope of DAL~C RSC development. Table~\ref{table_dal_C} includes the list of DAL~C objectives, their applicability by the DO-178C software levels and the compliance analysis column. This table complements the DAL~D objectives analysis presented in \cite{ML_DAL_D_case_study} and therefore doesn't include DAL~D objectives which are also applicable
to DAL~C. Since the generation of target source code from DNNs has no significant disparities compared to traditional software assurance practices covered by DO-178C, the source code development and verification objectives are not included in the analysis. Also, this analysis does not encompass the objectives related to planning, configuration management, quality assurance, and certification liaison processes. These processes are not significantly affected by ML-specific challenges and can be fulfilled using existing assurance practices \cite{ML_DAL_C_workflow}.
Overall 14 out of 23 DO-178C and DO-330 objectives included in the analyzed DAL~C subset have been directly accomplished, 4 objectives for executable code testing have been covered at the RSC DNN model level, and 5 objectives have been justified through the application of architectural mitigation. 

\section{Conclusions and Future Work}
\label{sec:conclusions} 
This paper focuses on the certification and safety aspects of machine-learning systems 
used in safety-critical airborne applications. To demonstrate the approach for addressing the certification issues of such systems, we conducted a case study of an onboard system utilizing deep neural networks for the detection and classification of airport runway signs. We demonstrated that machine learning systems can attain Design Assurance Level C by utilizing our custom development workflow in combination with an architectural mitigation approach and selected novel ML-specific data verification methods. Building upon our previous work that studied the Design Assurance Level D certification workflow, this case study involves the use of two dissimilar DNN components running in parallel. As part of the architectural mitigation approach, we integrated the safety monitor component, which continuously compares the outputs of dissimilar DNNs to detect potential errors in DNN outputs and prevent error propagation.

Dissimilarity of independent DNNs is achieved by using dissimilar network architectures, training frameworks, and data sets. On a more implementation-oriented level, we utilized different programming languages and having different team members doing the implementation. The distribution of detection divergence for dissimilar DNNs measured on the available dataset fits the normal distribution that correlates with the dissimilarity applied concepts. Based on the determined characteristics of the divergence distribution, we selected an optimal filtering threshold for the safety monitor component to ensure a target system availability of 95\%.

To verify compliance of the implemented system with Design Assurance Level C, we conducted a thorough analysis and justification of the DAL C objectives outlined in the DO-178C and DO-331 standards. Additionally, we mapped the analyzed objectives to the corresponding requirements specified in the latest ML-specific certification guidance published by EASA \cite{easaL1-2concept2023}.

In our future work, we will further study novel verification techniques specific to ML technologies with respect to certification objectives. In particular, we plan to focus on statistical data quality, model stability and robustness analysis methods, and out-of-distribution detection techniques. Additionally, we aim to explore the aspects of ML models' dissimilarity with respect to the Run Time Assurance approach described in ASTM 3269 \cite{ASTM.F3269}. Lastly, we intend to conduct the qualification of ML data traceability tools to streamline manual activities of data verification following the approach proposed in \cite{tool_qual_kd_2023}.

\noindent{\bf Acknowledgments.} We would like to thank  Shanza A. Zafar for the careful review of this paper and helpful feedback.

\bibliographystyle{IEEEtran}
\bibliography{bibliography}

\begin{table*}[t]
\caption{DAL C Objectives Analysis}
\label{table_dal_C}
\footnotesize
\begin{tabular}{|l|>{\raggedright}p{0.2\textwidth}|p{0.1mm}|p{0.1mm}|p{0.1mm}|p{0.1mm}|p{0.1mm}|p{0.55\textwidth}|}
\hline
\multicolumn{2}{|c|}{DO-178C/DO-331 Objectives}&\multicolumn{5}{c|}{Software Levels}&\multicolumn{1}{c|}{Compliance Analysis}\\
\multicolumn{2}{|c|}{}&A&B&C&D&E&\\
\hline\hline
A-2\#4&Low-level requirements are developed.
&x&x&x&&&
Trained RSC DNN models represent software design (low-level requirements and architecture); the objective is satisfied. This corresponds to the objective IMP-03 of the EASA Guidance \cite{easaL1-2concept2023}.\\
\hline
A-2\#5&Derived low-level requirements are defined and provided to the system processes, including the safety assessment process.\\
&x&x&x&&&
The \textit{traceability issue} discussed in \cite{ML_DAL_C_workflow} makes it practically infeasible to directly identify derived requirements associated with RSC DNNs. The EASA Guidance \cite{easaL1-2concept2023} does not have corresponding objectives. The justification for this objective relies on the implemented architectural mitigation, which includes utilizing two dissimilar RSC DNNs and a safety monitor.\\
\hline
A-3\#4&High-level requirements are verifiable.
&x&x&x&&&
The RSC system, component and data requirements written in textual form have been peer-reviewed to accomplish this objective. This corresponds to the objective DA-04 of the EASA Guidance \cite{easaL1-2concept2023}.\\
\hline
A-3\#5&High-level requirements conform to standards.
&x&x&x&&&
Same as for the objective A-3\#4.\\
\hline
A-3\#7&Algorithms are accurate. 
&i&i&x&&&
Same as for the objective A-3\#4.\\
\hline
A-4\#1&Low-level requirements comply with high-level requirements.
&i&i&x&&&
This objective is achieved through the testing of the RSC DNNs, which represent low-level software requirements in the implemented custom workflow. This corresponds to the combination of the objectives LM-09 and LM-10  of the EASA Guidance \cite{easaL1-2concept2023}.\\
\hline
A-4\#2&Low-level requirements are accurate and consistent.
&i&i&x&&&
Same as for the objective A-4\#1.\\
\hline
A-4\#5&Low-level requirements conform to standards.
&x&x&x&&&
This objective has been fulfilled by conducting peer reviews of the RSC DNN models representing software design. The EASA Guidance \cite{easaL1-2concept2023} has no corresponding objectives; this could indicate a potential gap in the provided guidance.\\
\hline
A-4\#6&Low-level requirements are traceable to high-level requirements.
&x&x&x&&&
Similar to objective A-2\#5, this objective is impacted by the \textit{traceability issue}. The EASA Guidance \cite{easaL1-2concept2023} does not include corresponding objectives. The implemented architectural mitigation measures support the justification for this objective.\\
\hline
A-4\#7&Algorithms are accurate.
&i&i&x&&&
Same as for the objective A-4\#1.\\
\hline
A-4\#8&Software architecture is compatible with high-level requirements.
&i&x&x&&&
Same as for the objective A-4\#1.\\
\hline
A-4\#9&Software architecture is consistent.
&i&x&x&&&
Same as for the objective A-4\#1.\\
\hline
A-4\#12&Software architecture conforms to standards.
&x&x&x&&&
Same as for the objective A-4\#5.\\
\hline
MB.A-4\#MB14&Simulation cases are correct
&i&x&x&&&
Since DNN model testing is used to fulfill the objectives for low-level requirements verification; it is necessary to verify the correctness of the DNN model test cases to ensure comprehensive testing. This objective is achieved through verification of requirements coverage by test data set as outlined in Section~\ref{sec:vnv}. This is partially addressed (for test dataset only) by the objective DM-14 the EASA Guidance \cite{easaL1-2concept2023}.\\
\hline
MB.A-4\#MB15&Simulation procedures are correct
&i&x&x&&&
Similar to the rationale behind objective MB.A-4\#MB14, it is necessary to verify the correctness of test procedures for the DNN model. RSC DNNs test procedures have been peer-reviewed to satisfy this objective. The EASA Guidance \cite{easaL1-2concept2023} has no corresponding objectives; this could indicate a potential gap in the provided guidance.\\
\hline
MB.A-4\#MB16&Simulation results are correct and discrepancies explained
&i&x&x&&&
Similar to the rationale behind objective MB.A-4\#MB14, it is necessary to verify DNN testing results. These testing results have been peer-reviewed to satisfy this objective. This is implicitly addressed by the objective LM-10 of the EASA Guidance \cite{easaL1-2concept2023}.\\
\hline
A-6\#3&Executable Object Code complies with low-level requirements.
&i&i&x&&&
To fulfill this objective, back-to-back testing of the target executable object code against the RSC DNN model can be utilized as described in \cite{ISO.26262}. Since the deployment of RSC DNN on target hardware was determined to have no significant disparities compared to traditional software assurance practices covered by DO-178C \cite{ML_DAL_C_workflow}, the testing objectives were implemented at the RSC DNN model level only, see A-4 and MB.A-4 objectives. This is covered by the objective IMP-03 of the EASA Guidance \cite{easaL1-2concept2023}.\\
\hline
A-6\#4&Executable Object Code is robust with low-level requirements.
&i&x&x&&&
Same as for the objective A-6\#3.\\
\hline
A-7\#1&Test procedures are correct.
&i&x&x&&&
Verification of test procedures through reviews can be utilized to satisfy this objective. This activity was replaced by MB.A-4\#MB15 in the scope of this case study, see objective A-6\#3.\\
\hline
A-7\#2&Test results are correct and discrepancies explained.
&i&x&x&&&
Verification of test results through reviews can be utilized to satisfy this objective.  This activity was replaced by MB.A-4\#MB16 in the scope of this case study, see objective A-6\#3.\\
\hline
A-7\#4&Test coverage of low-level requirements is achieved.
&i&x&x&&&
Due to the inherent \textit{coverage issue} discussed in \cite{ML_DAL_C_workflow}, there are no relevant metrics for the RSC DNN coverage analysis. The EASA Guidance \cite{easaL1-2concept2023} has no corresponding objectives. The justification for this objective is claimed from the implemented architectural mitigation, which involves the utilization of two dissimilar RSC DNNs and a safety monitor.\\
\hline
A-7\#7&Test coverage of software structure (statement coverage) is achieved.
&i&i&x&&&
As discussed in \cite{ML_DAL_C_workflow}, structural coverage metrics at the DNN source code level are not representative due to the inherent \textit{coverage issue } and have not been included in the scope of this work. The EASA Guidance \cite{easaL1-2concept2023} has no corresponding objectives. The justification for this objective is claimed from the implemented RSC architectural mitigation.\\
\hline
A-7\#8&Test coverage of software structure (data coupling and control coupling) is achieved.
&i&i&x&&&
Analysis of data and control coupling can be implemented using traditional methods to satisfy this objective. This is covered by the objective IMP-03 of the EASA Guidance \cite{easaL1-2concept2023}. The verification activities at the source code level have not been included in the scope of this case study, see the objective A-6\#2.\\
\hline
\end{tabular}
\end{table*}

\end{document}